# Developing Predictive and Robust Radiomics Models for Chemotherapy Response in High-Grade Serous Ovarian Carcinoma

Sepideh Hatamikia[†*, 1,2], Geevarghese George[†, 3], Florian Schwarzhans[3], Amirreza Mahbod[3], Marika AV Reinius[8,9,10], Ali Abbasian Ardakani[1], Mercedes Jimenez-Linan[8,11], Satish Viswanath[4], Mireia Crispin-Ortuzar[8,9], Lorena Escudero Sanchez[5], Evis Sala[6,7#], James D Brenton[8,9,10#], Ramona Woitek[#3]

[1]Research Center for Clinical AI-Research in Omics and Medical Data Science, Danube Private University, Krems, Austria
[2]Austrian Center for Medical Innovation and Technology (ACMIT), Wiener Neustadt, Austria
[3]Research Center for Medical Image Analysis and Artificial Intelligence (MIAAI), Department of Medicine, Danube Private University, Krems, Austria
[4] Pediatrics and Biomedical Engineering, Emory University, USA
[5] Department of Radiology, University of Cambridge, UK
[6]Fondazione Policlinico Universitario Agostino Gemelli IRCCS, Rome, Italy
[7]Università Cattolica del Sacro Cuore, Rome, Italy
[8] Cancer Research UK Cambridge Centre, Cambridge, UK
[9] Early Cancer Institute, University of Cambridge, Cambridge, UK
[10]Department of Oncology, University of Cambridge, Cambridge, United Kingdom
[11]Department of Pathology, Cambridge University Hospitals NHS Foundation Trust, Cambridge, UK

*Corresponding author
† Authors with equal contributions
#Authors with equal contributions

# Developing Predictive and Robust Radiomics Models for Chemotherapy Response in High-Grade Serous Ovarian Carcinoma

## Abstract

**Objectives:** High-grade serous ovarian carcinoma (HGSOC) is typically diagnosed at an advanced stage with extensive peritoneal metastases, making treatment challenging. Neoadjuvant chemotherapy (NACT) is often used to reduce tumor burden before surgery, but about 40% of patients show limited response. Radiomics, combined with machine learning (ML), offers a promising non-invasive method for predicting NACT response by analyzing computed tomography (CT) imaging data. This study aimed to improve response prediction in HGSOC patients undergoing NACT by integration different feature selection methods.

**Materials and methods:** A framework for selecting robust radiomics features was introduced by employing an automated randomisation algorithm to mimic inter-observer variability, ensuring a balance between feature robustness and prediction accuracy. Four response metrics were used: chemotherapy response score (CRS), RECIST, volume reduction (VolR), and diameter reduction (DiaR). Lesions in different anatomical sites were studied. Pre- and post-NACT CT scans were used for feature extraction and model training on one cohort, and an independent cohort was used for external testing.

**Results:** The best prediction performance was achieved using all lesions combined for VolR prediction, with an AUC of 0.83. Omental lesions provided the best results for CRS prediction (AUC 0.77), while pelvic lesions performed best for DiaR (AUC 0.76).

**Conclusion:** The integration of robustness into the feature selection processes ensures the development of reliable models and thus facilitates the implementation of the radiomics models in clinical applications for HGSOC patients. Future work should explore further applications of radiomics in ovarian cancer, particularly in real-time clinical settings.

**Key Points**

- Integrating segmentation-robustness into radiomics feature selection helps maintain feature stability despite contour variability, enhancing the reliability, reproducibility, and clinical applicability of radiomics models for predicting chemotherapy response in high-grade serous ovarian carcinoma.
- Radiomics-based machine learning models demonstrated strong external validation performance, achieving an AUC of 0.83 for volume-based response prediction when features from all lesions were combined, highlighting the value of multi-site tumour characterization.
- Prediction performance varied substantially across anatomical sites and response metrics, with omental radiomics best predicting CRS and pelvic radiomics best predicting diameter-based response, underscoring the importance of site-specific modelling strategies in HGSOC.

**Clinical Relevance Statement**

Incorporating segmentation-robustness into radiomics feature selection enhances the stability and reliability of predictive models, enabling more consistent identification of patients unlikely to benefit from neoadjuvant chemotherapy and supporting more informed, individualized treatment planning for high-grade serous ovarian carcinoma in clinical practice.

# Introduction

High-grade serous ovarian carcinoma (HGSOC) typically presents at an advanced stage with multi-site peritoneal metastases and therefore poses a major therapeutic challenge for clinicians. A large proportion of patients undergo neoadjuvant chemotherapy (NACT) to reduce tumor burden before surgery, followed by delayed primary surgery (DPS) [1]. Although many patients show a favorable response to treatment, NACT is not objectively beneficial for approximately 40% of patients [2-4]. Identifying patients who are unlikely to benefit from NACT before treatment begins could improve clinical decision-making and help guide them towards immediate primary surgery or enrollment in trials of novel targeted therapies.

Several studies have addressed risk stratification and response prediction in HGSOC using imaging features and artificial intelligence [5-7]. Radiological imaging offers the advantage of non-invasively capturing the spatial heterogeneity of HGSOC and might thus help address the complexity of HGSOC at multiple scales, ranging from regions of visibly different attenuation on computed tomography (CT) to temporally and spatially heterogeneous tumour genomic features [5, 8, 9]. HGSOC radiomics in combination with machine learning (ML) have been used in several studies [7] although most have focused on ovarian/pelvic lesions and only rarely included whole tumour or omental radiomics [5, 6, 10]. Besides, some studies demonstrated that rim-based radiomics features [11] and habitat radiomics analysis [12] carry clinically meaningful information and can contribute to predicting chemotherapy treatment outcomes.

The datasets used in this study have previously supported the development of radiomics-based ML algorithms for predicting treatment response in HGSOC [5, 10]. Despite promising results, the potential to further enhance prediction performance by refining or adding elements within the radiomics workflow remains insufficiently explored. Lesion segmentation is a crucial step in radiomics studies, and it has been highlighted that segmentation variability profoundly affects robustness and repeatability of quantitative imaging features, including radiomics [13]. One way to addressing this challenge is to select radiomics features that remain robust despite variations in tumour delineation. In this work, we provide a framework for selecting robust radiomics features by applying an automated randomisation algorithm that introduces minor variations into segmentations while balancing feature robustness and prediction accuracy.

# Materials and Methods

## Patients and CT scans

The institutional review boards approved two prospective studies in patients with HGSOC undergoing NACT and CT before and after NACT (REC reference number 08/H0306/61 and IRAS reference number 243824). Written informed consent was obtained from all patients. Patients receiving NACT for HGSOC from the Cambridge University Hospitals NHS Foundation Trust between 2009 and 2017 were recruited as the training cohort (OV04). In addition, patients from the Barts Health NHS Trust were recruited between October 2009 and October 2014 and considered as the external test cohort (Barts). Clinical and outcome data from patients in the external test cohort were reported in a previous publication [12], and imaging data together with clinical and outcome data were reported for both cohorts in previous publications [5, 10].

Pre- and post-NACT abdominopelvic CT scans acquired after intravenous injection of iodinated contrast agent were included in this study. Details regarding inclusion criteria, selection criteria for NACT, patient management plans, and the collection of clinical and imaging data are described in [5, 10].

## Lesion segmentation

All cancer lesions were segmented semi-automatically using Microsoft Radiomics (InnerEye; Microsoft, Redmond, WA, USA) by either a board-certified radiologist with ten years of experience (RW) or a radiology resident whose segmentations were subsequently reviewed and adjusted by the same radiologist. Lesions were labelled by anatomic site, including omentum and ovaries/pelvis (=pelvic lesions).

## Histopathological CRS analysis

At both institutions, board-certified gynaecologic oncology pathologists assessed omental tumour specimens according to the three-tier CRS (CRS 1=abundant tumour with no or minimal perceptible response to chemotherapy; CRS 2=significant amount of viable tumour present, showing readily appreciable fibro-inflammatory response secondary to treatment; CRS 3=complete or near-complete response with no tumour or minimal irregularly scattered tumour nests (none>2mm)) [14, 15]. For analysis, CRS 1–2 were grouped as non-complete response and CRS 3 as complete response.

## RECIST analysis

A board-certified radiologist (RW) assessed the CT scans according to the RECIST 1.1 response criteria [5].

## Volume change

For volume analysis, we computed the percentage reduction of volume (VolR) between the pre- and post-NACT scans, with full details provided in the Supplementary Material. VolR was then binarized into response and non-response. Patients whose overall tumour volume decreased ≤65% (corresponding to ≤30% reduction in tumour diameter per RECIST 1.1) were considered having non-response, whereas patients with >65% volume reduction were considered having response [16].

## Diameter change

The reduction in tumour diameter (DiaR) between pre- and post-NACT CT scans was calculated for each patient using manually obtained RECIST 1.1 measurements (summed longest diameters (SLD)). As in RECIST 1.1., an SLD decrease ≤30% was considered non-response, whereas a decrease >30% was considered response [17].

## Analysis of omental, pelvic and all cancer lesions

We investigated the performance of radiomics-based predictors for CRS, RECIST, VolR, and DiaR using radiomics extracted from omental lesions only, pelvic lesions only, and all lesions combined. Training and testing of ML models were performed on the OV04 and BARTS cohort, respectively. The number of patients in each dataset ranged from 42 to 118, depending on the availability of scans and response labels and the suitability of scans and ROIs for processing (details in the Supplementary Material). We also investigated how using radiomics features from the largest lesion only ('Largest' approach) or from all lesions ('Merged' approach) affected the prediction results (Figure 1).

To mimic the variability in lesion delineation across and within radiologists, minor random changes were made to the contours of the original VOIs. Radiomics features were then extracted from both the unaltered and the changed VOIs. To assess the robustness of these features to VOI modifications, the intraclass correlation coefficient (ICC) was calculated, enabling the identification of radiomics features that remain consistent and reliable despite contour variation.

## Machine learning models and feature selection

Feature selection (FS) is essential in radiomics-based studies because of the high-dimensional feature space. In our study, we used several filter, wrapper and embedded FS algorithms. Highly correlated or irrelevant features were first removed using Univariate Feature Selection (UFS) within a 5-fold stratified cross-validation loop. This reduced feature set was then fed to our main FS algorithms: F-score, Relief, mutual information (MI), Gini importance (GI), least absolute shrinkage and selection operator (LASSO), genetic algorithm (GA), sequential backward search (SBS), sequential forward search (SFS), and recursive feature elimination (RFE). Logistic regression (LR) and linear discriminant analysis (LDA) served as the ML models. The best-performing combination of feature set and model was selected based on the area under the ROC curve (AUC) in the training dataset and subsequently evaluated on the test dataset to assess performance and generalizability. G-Mean, sensitivity (SE), and specificity (SP) were also reported.

## Randomisation of lesion segmentations

Manual lesion delineation by experts is considered the gold-standard, yet it remains subjective and prone to inter- and intra-operator variability, which can adversely affect radiomics by altering extracted features and leading to inconsistent model predictions [18]. In this study, small random modifications to the original VOIs were introduced in order to mimic multiple readers (Figure 1 and Figure 2 a,b; calculation details in the Supplementary Material). To quantify the robustness of individual radiomics features against changes in the segmentation, we calculated the ICC between randomized and original segmentations. These feature-wise ICC scores were then used in the feature selection pipelines (Figure 3). The randomisation was applied only to the training dataset for ICC computation.

## Rim radiomics features

It has been demonstrated that radiomics computed for the peripheral regions of tumours can provide valuable prognostic information [19, 20]. To extract rim radiomics features, a 6-mm-wide rim along the outline of VOIs was generated, including both the lesion periphery and a perilesional area (Figure 2c). Details of calculations can be found in Supplementary Material. Rim radiomics were calculated only for the single largest lesion overall or per anatomic location (omentum and pelvis) to prevent calculated rims from overlapping and merging in the case of neighbouring lesions. Rim radiomics features were used in models for CRS prediction only.

## Radiomics feature extraction

A total of 102 radiomics features, including first-order, shape-based, and texture-based features, were extracted from the segmented cancer lesions on CT scans using the open-source Python package Pyradiomics (version 3.0.1) in Python (version 3.9.13). A fixed bin-width of four was determined to be ideal [21].

## Selection of predictive and robust features

Many radiomics features have shown limited robustness to variations in lesion delineation and are therefore highly reader-dependent. To enable clinical translation of radiomics-based models, the prediction models developed must be robust to such variability. This can be achieved by directly selecting radiomics features based on their robustness and reproducibility. We propose three approaches that consider the mutual impact of predictive power and robustness (PREDICTIVE&ROBUST approaches) of radiomics features and compare them with an approach relying solely on predictive features (PREDICITVE approach). ICC values of radiomics features are used to identify features that are simultaneously robust and predictive. The three PREDICTIVE&ROBUST approaches are described below.

### *Fully robust*

In the fully robust approach, an ICC threshold of >0.8 was applied as a pre-step for feature selection to ensure that only highly robust features were included in the initial feature pool.

### *Semi-robust*

The semi-robust approach focused on the proportion of features with ICC scores >0.8 in the candidate feature pool during the feature selection step. At each iteration of the feature selection process, at least 80% of the features in the candidate pool needed to have ICC scores >0.8 to ensure that the majority of features being considered were robust.

### *Weighted robustness*

In the weighted robustness approach, the ICC score directly interacts with the feature relevance in a weighted manner during the feature selection step. At each iteration, the ICC scores of the features in the candidate pool are considered as,

$$s = (1 - w) * s + w * \bar{c}$$

Here, $w$ is the weighting term for the mean robustness, $\bar{c}$ measured by the ICC scores of the features under consideration, and s is the feature relevance score or statistic that dictates the feature selection algorithm. The modified feature selection step can prioritise the robustness and the relevance or predictive capacity of the feature set based on the parameter $w$. In our experiments, $w$ was set to 0.5 for equal weighting.

## Results

### Assessment of prediction results using PREDICTIVE approach

The best prediction results using the PREDICTIVE approach for CRS, RECIST, VolR, and DiaR based on omental, pelvic, and all lesions for both the training and test cohort are shown in Table 1. The best-performing FS method, best ML model, best feature extraction method (largest or merged), best classifier performance when no feature selection was applied, and number of selected features along with their names are reported (Table 1 and Supplementary Figure 1).

The features computed for all sites (merged) gave the best-performing model for VoIR prediction (AUC=0.83; G-mean=64, SE=0.85; SP=0.48) using SFS feature selection, an LDA classifier, and merging of all lesions. The features extracted from the omental site gave the best-performing model for CRS prediction (AUC=0.75; G-Mean=59, SE=0.98; SP=0.35) using SFS feature selection, an LR classifier, and merging of all lesions, whereas features extracted from the pelvic site gave the best-performing model for DiaR prediction (AUC=0.75; G-Mean=72, SE=0.66; SP=0.79) using F-Score feature selection, an LDA classifier, and merging of all lesions. The best-performing feature selection algorithms, shown in Table 1, can reduce the feature space significantly while maintaining or improving prediction performance. We found that for different response metrics and also for different anatomical sites (omentum, pelvis, and all lesions), prediction performance differed, and different feature selection algorithms performed best, with different features selected per approach (Table 1, **Error! Reference source not found.**).

## Radiomics feature robustness analysis

The ICC values between the randomised and manually drawn segmentations for all features together and for each feature individually are shown in Figure 4. In addition, the percentage of features with ICC values above 0.9 (rated as excellent robustness) for all features and for different feature groups (shape, first order, texture) is shown in Figure 4. ICC values were above 0.9 for all shape features and for up to 68% of texture features, but only for 50% of first-order features.

## The PREDICTIVE&ROBUST approach

The best prediction results for the three PREDICITVE&ROBUST approaches (fully robust, semi-robust and weighted robustness) are shown in Tables 2-5. The selected features for the best-performing models using these three approaches can be found in the Supplementary Material. A comparison between the selected features using the PREDICTIVE approach (Figure 5) and models based on the PREDICITVE&ROBUST approach (Supplementary Figure 1 and Supplementary Figure 2) shows that different features were selected in the two approaches. Using lesions from all sites, the best model selected using the PREDICITVE&ROBUST approach was for VoIR prediction (similar to the PREDICTIVE approach), and similar performance to the PREDICTIVE approach was obtained. In addition, when using lesions from the omental site, the best-performing model was again for CRS prediction (AUC=0.77; G-Mean=72, SE=0.56; SP=0.92) with SFS as the best-performing feature selection method. However, when lesions from the pelvic region were used, the best model for PREDICITVE&ROBUST approach was selected for DiaR prediction (AUC=0.76; G-Mean=34, SE=0.12; SP=0.97), with SFS feature selection method. An increase in the average ICC value was achieved when using lesions from all anatomical sites as well as omental lesions for the PREDICITVE&ROBUST approach compared to the PREDICITVE approach (Tables 1-5). The distribution of feature robustness for the PREDICITVE&ROBUST compared to the PREDICTIVE approach for the best-performing models in all cases is presented in Figure 5.

## Prediction of CRS using rim radiomics features

The best prediction results using the PREDICTIVE approach for the CRS metric based on rim radiomics features can be found in Supplementary Table 2. The model with radiomics features from the entire VOIs outlined by radiologists and rim radiomics features achieved very similar predictive performance for omental lesions (AUC=0.74 for normal radiomics features compared with AUC=0.74 for rim radiomics features). In contrast, the AUC dropped substantially when instead of features extracted from the entire VOI, only rim features were extracted from pelvic lesions (AUC=0.68 and AUC=0.42, respectively).

# Discussion

In this study, we developed radiomics models for predicting CRS, RECIST, VolR, and DiaR metrics based on CT data from the omental, pelvis, and all lesions of patients with HGSOC. We propose a framework that account for the robustness of radiomics features to variations in tumour delineation that are known to occur when multiple readers outline lesions. With our randomisation approach, we introduce variations in tumour delineation automatically and thus simulate a setting with multiple readers. The manual or semi-automated segmentation of lesions for radiomics studies by a single expert, let alone several, is frequently a bottleneck hampering the use of large datasets in AI studies. Our approach may help overcome this shortage of imaging experts by generating sets of segmentations with slight variations based on an initial set of manually or semi-automatically generated segmentations.

We found that the performance of the model and the features selected were highly dependent on the type of response metric and the anatomic site from which the features were extracted. Features from all lesions resulted in the best-performing model (with an AUC of up to 0.83 for VolR prediction). For omental and pelvic lesions, similar AUC values (0.77, 0.76 respectively) were obtained with the best models, but a better balance between SE and SP was seen with the model for omental lesions compared to pelvic lesions. The use of RECIST as a response metric did not perform well (maximum AUC of 0.65 using omental lesions). In addition, using DiaR as a response metric yielded the best model when using pelvic lesions with semi-weighted approach (AUC=0.76), but a large difference was observed between SE and SP. The fully robust and weighted methods, however, performed better, with a slightly lower AUC of 0.75.

The best performing models selected under the PREDICTIVE&ROBUST achieved prediction performance that matched or exceeded that of the predictive PREDICTIVE approach across all cases. Given the constraints imposed by the PREDICTIVE&ROBUST approach to ensure both mutual predictive performance and robustness, the resulting outcomes are encouraging. Notably, we observed differences between the selected features using PREDICTIVE&ROBUST and PREDICTIVE approaches, highlighting that pure predictive features were not always the robust ones.

Compared to related studies aiming to predict NACT in patients with advanced ovarian cancer, our approach has the advantage of using an external dataset for validation of the developed model [22]. Moreover, our results are interpretable compared to the black-box nature of the deep learning methods developed by Yin et. al. [23] and we achieved a 16% increase in AUC in the external cohort compared with the work of Rundo et. al. [10]. On the other hand, Huang et al. [11] demonstrated that rim-based radiomics features carry clinically meaningful information and can contribute to predicting treatment outcomes, which aligns with our observations. Similarly, Bi et al. [12] introduced a habitat-radiomics framework using multi-parametric MRI and underscored its potential for forecasting response to chemotherapy. Future research should extend this approach to CT imaging to determine whether habitat-level characterization can likewise enhance prediction of chemotherapy response in patients with HGSOC.

In addition, in our study, we specifically investigated prediction performance across different anatomical sides and different prediction metrics, and we proposed a new framework to incorporate robustness of radiomics features into the feature selection process, addressing considerations that were unexplored in previous related research [10, 22, 23]. Integration of observer variability using a robustness score in the FS process and together with the use of an unseen patient cohort for validation, ensures the reliability and robustness of the proposed radiomics models. We also observed similar predictive performance between the two approaches using normal and rim radiomics features for omental lesions, while no meaningful prediction could be achieved with rim features extracted from pelvic lesions.

In conclusion, our study successfully developed robust radiomics models for predicting multiple response metrics in ovarian cancer. Compared to previous studies, our approach achieved significant improvements in predictive accuracy and robustness by incorporating observer variability into the feature selection process, supporting the development of robust radiomics models for HGSOC patients.

**Table 1:** Prediction results related to all lesion types, all prediction metrics for both train and test datasets as well as number of selected features and average ICC values. To quantify the change in performance, we define change w.r.t. no FS (%) as the difference between test dataset results with and without feature selection divided by test dataset results without feature selection. AUC: area under the ROC curve, FSA: feature selection algorithm, SE: sensitivity, SP: specificity, NF: number of features, ICC: intraclass correlation coefficient.

| Response | Lesions, Dataset, Model | FSA | Train (OV04) | | | | Hidden (BARTS) | | | | Change (%) | NF | Avg ICC |
|---|---|---|---|---|---|---|---|---|---|---|---|---|---|
| | | | AUC | GMean | SE | SP | AUC | GMean | SE | SP | | | |
| CRS | All, Largest, LDA | No | 0.51 | 0.52 | 0.44 | 0.61 | 0.54 | 0.41 | 0.22 | 0.75 | - | 102 | 0.81 (0.27) |
| CRS | All, Largest, LDA | F-Score | 0.62 | 0.46 | 0.24 | 0.89 | 0.73 | 0.37 | 0.14 | 0.96 | 19, -4, -8, 21 | 1 | 0.99 (0.0) |
| CRS | Omental, Merged, LR | No | 0.69 | 0.59 | 0.6 | 0.59 | 0.63 | 0.63 | 0.68 | 0.59 | - | 102 | 0.81 (0.27) |
| **CRS** | **Omental, Merged, LR** | **SFS** | **0.73** | **0.7** | **0.88** | **0.56** | **0.75** | **0.59** | **0.98** | **0.35** | **12, -4, 30, -24** | **2** | **0.86 (0.0)** |
| CRS | Pelvic, Largest, LDA | No | 0.48 | 0.47 | 0.37 | 0.59 | 0.57 | 0.5 | 0.38 | 0.67 | - | 102 | 0.81 (0.27) |
| CRS | Pelvic, Largest, LDA | F-Score | 0.63 | 0.39 | 0.17 | 0.88 | 0.68 | 0.55 | 0.36 | 0.84 | 11, 5, -2, 17 | 1 | 0.73 (0.0) |
| RECIST | All, Merged, LDA | No | 0.69 | 0.62 | 0.77 | 0.5 | 0.6 | 0.51 | 0.75 | 0.35 | - | 102 | 0.81 (0.27) |
| RECIST | All, Merged, LDA | F-Score | 0.74 | 0.63 | 0.79 | 0.5 | 0.61 | 0.49 | 0.74 | 0.32 | 1, -2, -1, -3 | 23 | 0.76 (0.27) |
| RECIST | Omental, Merged, LR | No | 0.61 | 0.6 | 0.64 | 0.56 | 0.64 | 0.63 | 0.74 | 0.54 | - | 102 | 0.81 (0.27) |
| RECIST | Omental, Merged, LR | LASSO | 0.72 | 0.64 | 0.71 | 0.58 | 0.65 | 0.44 | 0.85 | 0.23 | 1, -19, 11, -31 | 4 | 0.96 (0.06) |
| RECIST | Pelvic, Merged, LR | No | 0.53 | 0.55 | 0.7 | 0.43 | 0.57 | 0.53 | 0.74 | 0.38 | - | 102 | 0.81 (0.27) |
| RECIST | Pelvic, Merged, LR | ULR | 0.63 | 0.56 | 0.73 | 0.43 | 0.57 | 0.54 | 0.72 | 0.4 | 0, 1, -2, 2 | 19 | 0.71 (0.31) |
| VoIR | All, Merged, LDA | No | 0.65 | 0.64 | 0.62 | 0.67 | 0.75 | 0.64 | 0.81 | 0.51 | - | 102 | 0.81 (0.27) |
| **VoIR** | **All, Merged, LDA** | **SFS** | **0.76** | **0.71** | **0.82** | **0.62** | **0.83** | **0.64** | **0.85** | **0.48** | **8, 0, 4, -3** | **3** | **0.96 (0.03)** |
| VoIR | Omental, Merged, LR | No | 0.61 | 0.58 | 0.75 | 0.45 | 0.68 | 0.43 | 0.83 | 0.22 | - | 102 | 0.81 (0.27) |
| VoIR | Omental, Merged, LR | SBS | 0.76 | 0.74 | 0.81 | 0.67 | 0.78 | 0.51 | 0.91 | 0.29 | 10, 8, 8, 7 | 6 | 0.8 (0.28) |
| VoIR | Pelvic, Merged, LDA | No | 0.6 | 0.53 | 0.42 | 0.67 | 0.7 | 0.52 | 0.33 | 0.83 | - | 102 | 0.81 (0.27) |
| VoIR | Pelvic, Merged, LDA | ULR | 0.71 | 0.56 | 0.44 | 0.72 | 0.74 | 0.58 | 0.36 | 0.92 | 4, 6, 3, 9 | 28 | 0.73 (0.29) |
| DiaR | All, Largest, LDA | No | 0.64 | 0.62 | 0.75 | 0.51 | 0.52 | 0.48 | 0.73 | 0.31 | - | 102 | 0.81 (0.27) |
| DiaR | All, Largest, LDA | SBS | 0.72 | 0.68 | 0.71 | 0.65 | 0.54 | 0.55 | 0.59 | 0.52 | 2, 7, -14, 21 | 9 | 0.76 (0.27) |
| DiaR | Omental, Merged, LDA | No | 0.54 | 0.47 | 0.75 | 0.3 | 0.65 | 0.61 | 0.67 | 0.55 | - | 102 | 0.81 (0.27) |
| DiaR | Omental, Merged, LDA | GA | 0.71 | 0.73 | 0.78 | 0.68 | 0.67 | 0.63 | 0.62 | 0.65 | 2, 2, -5, 10 | 9 | 0.88 (0.2) |
| DiaR | Pelvic, Largest, LDA | No | 0.62 | 0.59 | 0.48 | 0.72 | 0.66 | 0.64 | 0.68 | 0.6 | - | 102 | 0.81 (0.27) |

| DiaR | Pelvic, Largest, LDA | F-Score | 0.73 | 0.62 | 0.54 | 0.72 | 0.75 | 0.72 | 0.66 | 0.79 | 9, 8, -2, 19 | 6 | 0.96 (0.02) |

**Table 2:** CRS prediction results related to all lesion types and three robustness approaches for both train and test datasets as well as number of selected features and average ICC values. To quantify the change in performance, we define change w.r.t. no FS (%) as the difference between test dataset results with and without feature selection divided by test dataset results without feature selection. FSA: feature selection algorithm, AUC: area under the ROC curve, SE: sensitivity, SP: specificity, NF: number of features, ICC: intraclass correlation coefficient.

| Lesion | Robustness Method | FSA | Train (OV04) | | | | Hidden (BARTS) | | | | Change (%) | NF | Avg ICC |
|---|---|---|---|---|---|---|---|---|---|---|---|---|---|
| | | | AUC | GMean | SE | SP | AUC | GMean | SE | SP | | | |
| All, Largest, LDA | No | No | 0.51 | 0.52 | 0.44 | 0.61 | 0.54 | 0.41 | 0.22 | 0.75 | - | 102 | 0.81 (0.27) |
| **All, Largest, LR** | **Full** | **SBS** | **0.71** | **0.63** | **0.6** | **0.66** | **0.79** | **0.68** | **0.52** | **0.88** | **25, 27, 30, 13** | **4** | **0.98 (0.01)** |
| All, Largest, LDA | Semi | SFS | 0.61 | 0.64 | 0.52 | 0.78 | 0.63 | 0.49 | 0.28 | 0.86 | 9, 8, 6, 11 | 10 | 0.89 (0.18) |
| All, Largest, LDA | Weighted | F-Score | 0.62 | 0.46 | 0.24 | 0.89 | 0.73 | 0.37 | 0.14 | 0.96 | 19, -4, -8, 21 | 1 | 0.99 (0.0) |
| Omental, Merged, LR | No | No | 0.69 | 0.59 | 0.6 | 0.59 | 0.63 | 0.63 | 0.68 | 0.59 | - | 102 | 0.81 (0.27) |
| **Omental, Largest, LR** | **Full** | **SFS** | **0.74** | **0.75** | **0.72** | **0.78** | **0.77** | **0.72** | **0.56** | **0.92** | **14, 9, -12, 33** | **2** | **0.93 (0.07)** |
| Omental, Merged, LR | Semi | SFS | 0.81 | 0.76 | 0.8 | 0.73 | 0.74 | 0.65 | 0.62 | 0.69 | 11, 2, -6, 10 | 10 | 0.87 (0.19) |
| Omental, Largest, LDA | Weighted | Gini | 0.77 | 0.59 | 0.4 | 0.86 | 0.77 | 0.53 | 0.3 | 0.94 | 14, -10, -38, 35 | 1 | 0.84 (0.0) |
| Pelvic, Largest, LDA | No | No | 0.48 | 0.47 | 0.37 | 0.59 | 0.57 | 0.5 | 0.38 | 0.67 | - | 102 | 0.81 (0.27) |
| Pelvic, Largest, LR | Full | SFS | 0.53 | 0.45 | 0.75 | 0.27 | 0.62 | 0.52 | 0.76 | 0.36 | 5, 2, 38, -31 | 3 | 0.96 (0.02) |
| Pelvic, Merged, LDA | Semi | RFE | 0.81 | 0.71 | 0.63 | 0.79 | 0.59 | 0.51 | 0.42 | 0.61 | 2, 1, 4, -6 | 11 | 0.94 (0.08) |
| Pelvic, Largest, LDA | Weighted | F-Score | 0.63 | 0.39 | 0.17 | 0.88 | 0.68 | 0.55 | 0.36 | 0.84 | 11, 5, -2, 17 | 1 | 0.73 (0.0) |

**Table 3: RECIST** prediction results related to all lesion types and three robustness approaches for both train and test datasets as well as number of selected features and average ICC values. To quantify the change in performance, we define change w.r.t. no FS (%) which calculate the difference between test dataset results with and without feature selection divided by test dataset results without feature selection. FSA: feature selection algorithm, AUC: area under the ROC curve, SE: sensitivity, SP: specificity, NF: number of features, ICC: intraclass correlation coefficient.

| Lesion | Robustness Method | FSA | Train (OV04) | | | | Hidden (BARTS) | | | | Change (%) | NF | Avg ICC |
|---|---|---|---|---|---|---|---|---|---|---|---|---|---|
| | | | AUC | GMean | SE | SP | AUC | GMean | SE | SP | | | |
| All, Merged, LDA | No | No | 0.69 | 0.62 | 0.77 | 0.5 | 0.6 | 0.51 | 0.75 | 0.35 | - | 102 | 0.81 (0.27) |
| All, Merged, LR | Full | SBS | 0.75 | 0.75 | 0.81 | 0.69 | 0.62 | 0.59 | 0.72 | 0.49 | 2, 8, -3, 14 | 7 | 0.96 (0.07) |
| All, Merged, LDA | Semi | ULR | 0.71 | 0.6 | 0.76 | 0.48 | 0.61 | 0.49 | 0.76 | 0.32 | 1, -2, 1, -3 | 32 | 0.8 (0.27) |
| All, Merged, LDA | Weighted | F-Score | 0.74 | 0.63 | 0.79 | 0.5 | 0.61 | 0.49 | 0.74 | 0.32 | 1, -2, -1, -3 | 23 | 0.76 (0.27) |
| Omental, Merged, LR | No | No | 0.61 | 0.6 | 0.64 | 0.56 | 0.64 | 0.63 | 0.74 | 0.54 | - | 102 | 0.81 (0.27) |
| Omental, Merged, LR | Full | LASSO | 0.69 | 0.65 | 0.7 | 0.61 | 0.64 | 0.47 | 0.86 | 0.26 | 0, -16, 12, -28 | 2 | 0.92 (0.06) |
| Omental, Merged, LR | **Semi** | **GA** | **0.68** | **0.67** | **0.65** | **0.7** | **0.65** | **0.61** | **0.75** | **0.5** | **1, -2, 1, -4** | **2** | **0.91 (0.06)** |
| Omental, Merged, LDA | Weighted | SFS | 0.68 | 0.6 | 0.73 | 0.49 | 0.65 | 0.37 | 0.89 | 0.15 | 1, -26, 15, -39 | 2 | 0.99 (0.01) |
| Pelvic, Merged, LR | No | No | 0.53 | 0.55 | 0.7 | 0.43 | 0.57 | 0.53 | 0.74 | 0.38 | - | 102 | 0.81 (0.27) |
| Pelvic, Merged, LR | Full | MI | 0.7 | 0.35 | 1 | 0.12 | 0.56 | 0.1 | 1 | 0.01 | -1, -43, 26, -37 | 1 | 0.99 (0.0) |
| Pelvic, Merged, LR | Semi | ULR | 0.63 | 0.56 | 0.73 | 0.43 | 0.57 | 0.54 | 0.72 | 0.4 | 0, 1, -2, 2 | 19 | 0.71 (0.31) |
| Pelvic, Merged, LR | Weighted | SBS | 0.62 | 0.6 | 0.69 | 0.52 | 0.57 | 0.58 | 0.68 | 0.49 | 0, 5, -6, 11 | 12 | 0.88 (0.24) |

**Table 4:** VolR prediction results related to all lesion types and three robustness approaches for both train and test datasets as well as number of selected features and average ICC values. To quantify the change in performance, we define Change w.r.t. no Fs (%) as the difference between test dataset results with and without feature selection divided by test dataset results without feature selection. FSA: feature selection algorithm, AUC: Area under the ROC Curve, SE: sensitivity, SP: specificity, Nf: number of features, ICC: Intraclass correlation coefficient.

| Lesion | Robustness Method | FSA | Train (OV04) | | | | Hidden (BARTS) | | | | Change (%) | NF | Avg ICC |
|---|---|---|---|---|---|---|---|---|---|---|---|---|---|
| | | | AUC | GMean | SE | SP | AUC | GMean | SE | SP | | | |
| All, Merged, LDA | No | No | 0.65 | 0.64 | 0.62 | 0.67 | 0.75 | 0.64 | 0.81 | 0.51 | - | 102 | 0.81 (0.27) |
| All, Merged, LDA | **Full** | **GA** | **0.76** | **0.71** | **0.82** | **0.62** | **0.83** | **0.64** | **0.85** | **0.48** | **8, 0, 4, -3** | **3** | **0.96 (0.03)** |
| All, Merged, LDA | **Semi** | **SFS** | **0.76** | **0.71** | **0.82** | **0.62** | **0.83** | **0.64** | **0.85** | **0.48** | **8, 0, 4, -3** | **3** | **0.96 (0.03)** |
| All, Merged, LDA | Weighted | LASSO | 0.74 | 0.69 | 0.76 | 0.63 | 0.79 | 0.67 | 0.81 | 0.55 | 4, 3, 0, 4 | 13 | 0.88 (0.22) |
| Omental, Merged, LR | No | No | 0.61 | 0.58 | 0.75 | 0.45 | 0.68 | 0.43 | 0.83 | 0.22 | - | 102 | 0.81 (0.27) |
| Omental, Merged, LDA | Full | LASSO | 0.77 | 0.55 | 0.93 | 0.32 | 0.76 | 0.38 | 0.95 | 0.15 | 8, -5, 12, -7 | 6 | 0.93 (0.06) |
| Omental, Merged, LDA | Semi | RFE | 0.78 | 0.55 | 0.96 | 0.32 | 0.75 | 0.33 | 0.97 | 0.11 | 7, -10, 14, -11 | 8 | 0.86 (0.18) |
| Omental, Merged, LR | Weighted | SFS | 0.71 | 0.65 | 0.77 | 0.55 | 0.77 | 0.41 | 0.95 | 0.18 | 9, -2, 12, -4 | 1 | 1.0 (0.0) |
| Pelvic, Merged, LDA | No | No | 0.6 | 0.53 | 0.42 | 0.67 | 0.7 | 0.52 | 0.33 | 0.83 | - | 102 | 0.81 (0.27) |
| Pelvic, Largest, LR | Full | Gini | 0.61 | 0.6 | 0.54 | 0.66 | 0.73 | 0.64 | 0.83 | 0.5 | 3, 12, 50, -33 | 2 | 0.98 (0.01) |
| Pelvic, Merged, LDA | Semi | ULR | 0.71 | 0.56 | 0.44 | 0.72 | 0.74 | 0.58 | 0.36 | 0.92 | 4, 6, 3, 9 | 28 | 0.73 (0.29) |
| Pelvic, Merged, LDA | Weighted | ULR | 0.71 | 0.56 | 0.44 | 0.72 | 0.74 | 0.58 | 0.36 | 0.92 | 4, 6, 3, 9 | 28 | 0.73 (0.29) |

**Table 5:** DiaR prediction results related to all lesion types and three robustness approaches for both train and test datasets as well as number of selected features and average ICC values. To quantify the change in performance, we define Change w.r.t. no Fs (%) as the difference between test dataset results with and without feature selection divided by test dataset results without feature selection. FSA: feature selection algorithm, AUC: Area under the ROC Curve, SE: sensitivity, SP: specificity, Nf: number of features, ICC: Intraclass correlation coefficient.

| Lesion | Robustness Method | FSA | Train (OV04) | | | | Hidden (BARTS) | | | | Change (%) | NF | Avg ICC |
| --- | --- | --- | --- | --- | --- | --- | --- | --- | --- | --- | --- | --- | --- |
| | | | AUC | GMean | SE | SP | AUC | GMean | SE | SP | | | |
| All, Largest, LDA | No | No | 0.64 | 0.62 | 0.75 | 0.51 | 0.52 | 0.48 | 0.73 | 0.31 | - | 102 | 0.81 (0.27) |
| All, Largest, LR | Full | RFE | 0.7 | 0.67 | 0.7 | 0.64 | 0.54 | 0.53 | 0.59 | 0.47 | 2, 5, -14, 16 | 8 | 0.96 (0.06) |
| All, Largest, LDA | Semi | ULR | 0.7 | 0.65 | 0.68 | 0.62 | 0.54 | 0.53 | 0.59 | 0.48 | 2, 5, -14, 17 | 18 | 0.75 (0.3) |
| All, Largest, LDA | Weighted | ULR | 0.7 | 0.65 | 0.68 | 0.62 | 0.54 | 0.53 | 0.59 | 0.48 | 2, 5, -14, 17 | 18 | 0.75 (0.3) |
| Omental, Merged, LDA | No | No | 0.54 | 0.47 | 0.75 | 0.3 | 0.65 | 0.61 | 0.67 | 0.55 | - | 102 | 0.81 (0.27) |
| Omental, Merged, LDA | Full | SBS | 0.72 | 0.71 | 0.85 | 0.6 | 0.67 | 0.65 | 0.66 | 0.64 | 2, 4, -1, 9 | 6 | 0.92 (0.05) |
| Omental, Merged, LDA | Semi | GA | 0.72 | 0.71 | 0.82 | 0.62 | 0.68 | 0.65 | 0.65 | 0.65 | 3, 4, -2, 10 | 4 | 0.92 (0.06) |
| Omental, Merged, LDA | Weighted | GA | 0.71 | 0.68 | 0.82 | 0.57 | 0.66 | 0.65 | 0.65 | 0.65 | 1, 4, -2, 10 | 6 | 0.95 (0.06) |
| Pelvic, Largest, LDA | No | No | 0.62 | 0.59 | 0.48 | 0.72 | 0.66 | 0.64 | 0.68 | 0.6 | - | 102 | 0.81 (0.27) |
| Pelvic, Largest, LDA | **Full** | **SFS** | **0.7** | **0.64** | **0.52** | **0.78** | **0.75** | **0.72** | **0.64** | **0.82** | 9, 8, -4, 22 | **4** | **0.97 (0.01)** |
| Pelvic, Merged, LDA | **Semi** | **GA** | **0.76** | **0.61** | **0.39** | **0.96** | **0.76** | **0.34** | **0.12** | **0.97** | 10, -30, -56, 37 | **10** | **0.94 (0.08)** |
| Pelvic, Largest, LDA | **Weighted** | **SFS** | **0.71** | **0.66** | **0.57** | **0.76** | **0.75** | **0.74** | **0.68** | **0.8** | 9, 10, 0, 20 | **5** | **0.98 (0.01)** |

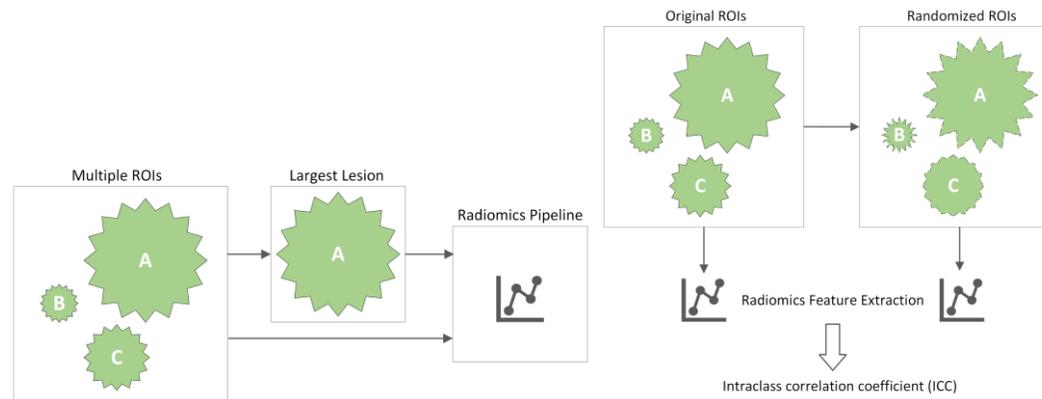

**Figure 1:** Four approaches for radiomics computation and integration in model building. Either (a) the largest lesion selected from multiple OIs and radiomics features are extracted, or (b) radiomics features are extracted from all VOIs for each CT scan.

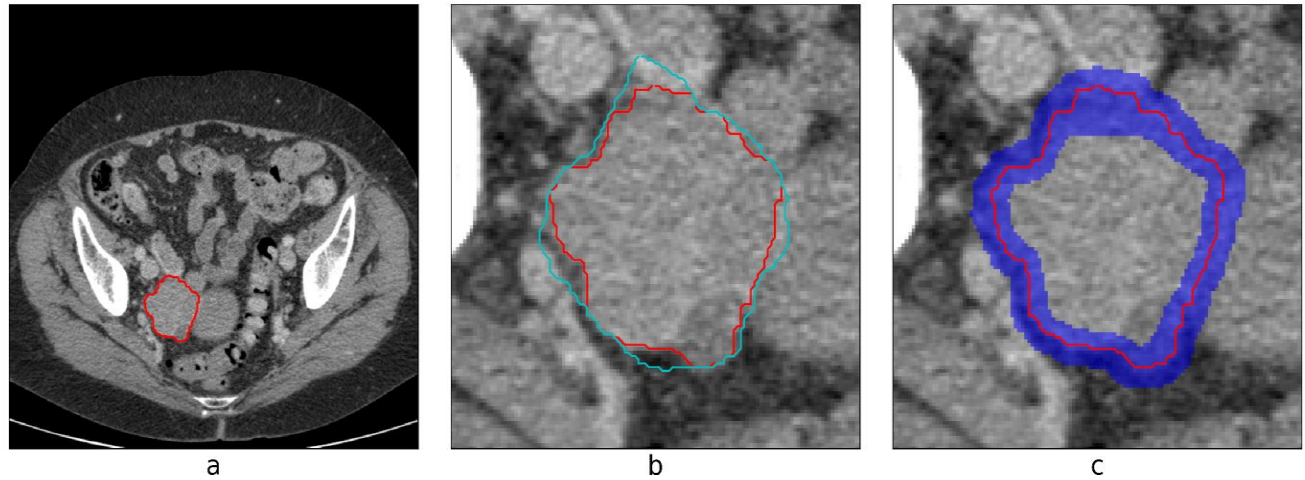

**Figure 2:** Example of a CT scan with (a) original volume of interest (VOI) (red) generated by a radiologist, (b) VOI (cyan) with random variations introduced to the radiologist's segmentation and (c) a 6 mm rim VOI (blue) including the periphery of the lesion and a perilesional area.

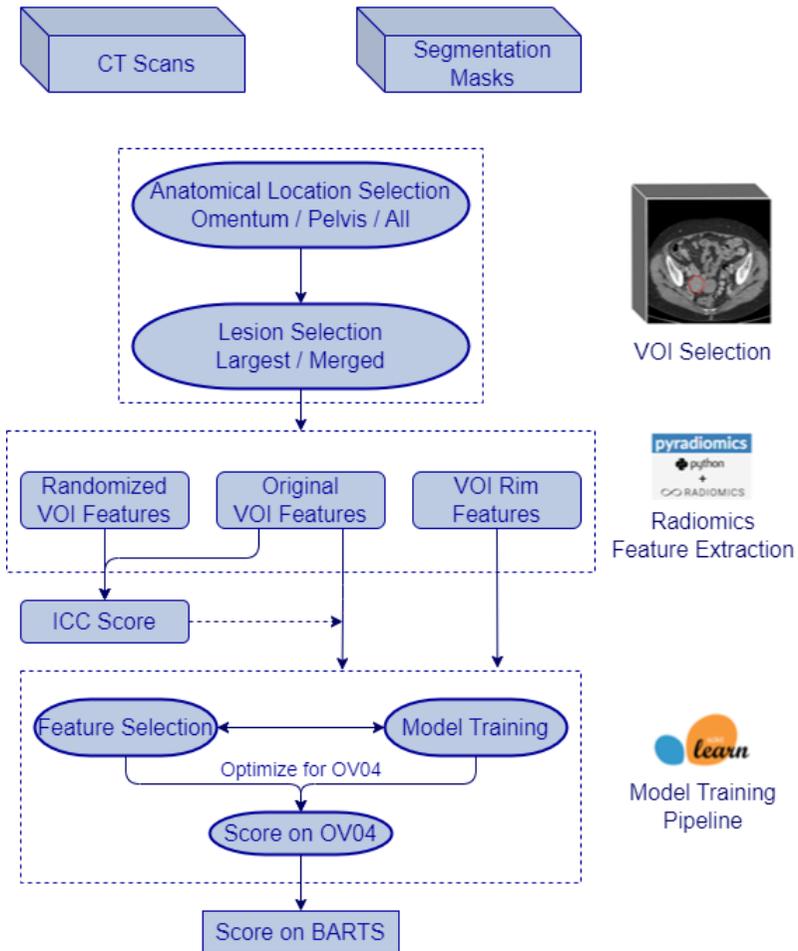

**Figure 3:** Overview diagram illustrating the sequential stages of our workflow for chemotherapy response prediction.

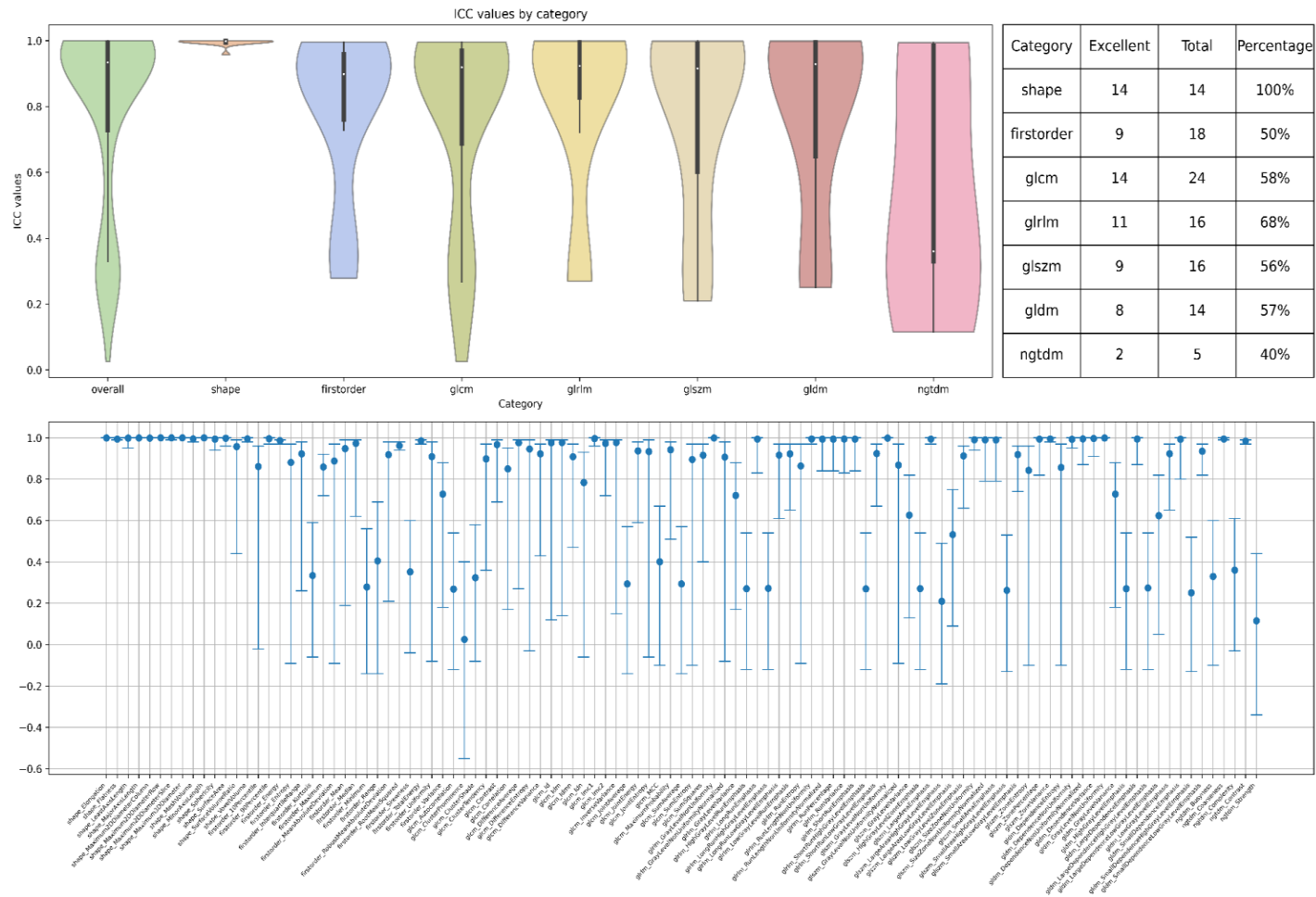

**Figure 4:** ICC values for different radiomics features and feature categories. (a) Distribution of ICC values separated by feature category; (b) Percentage of excellent features with an ICC >= 0.9 (features rated as excellent) per category; (c) ICC value including 95% confidence interval for each radiomics feature (bottom).

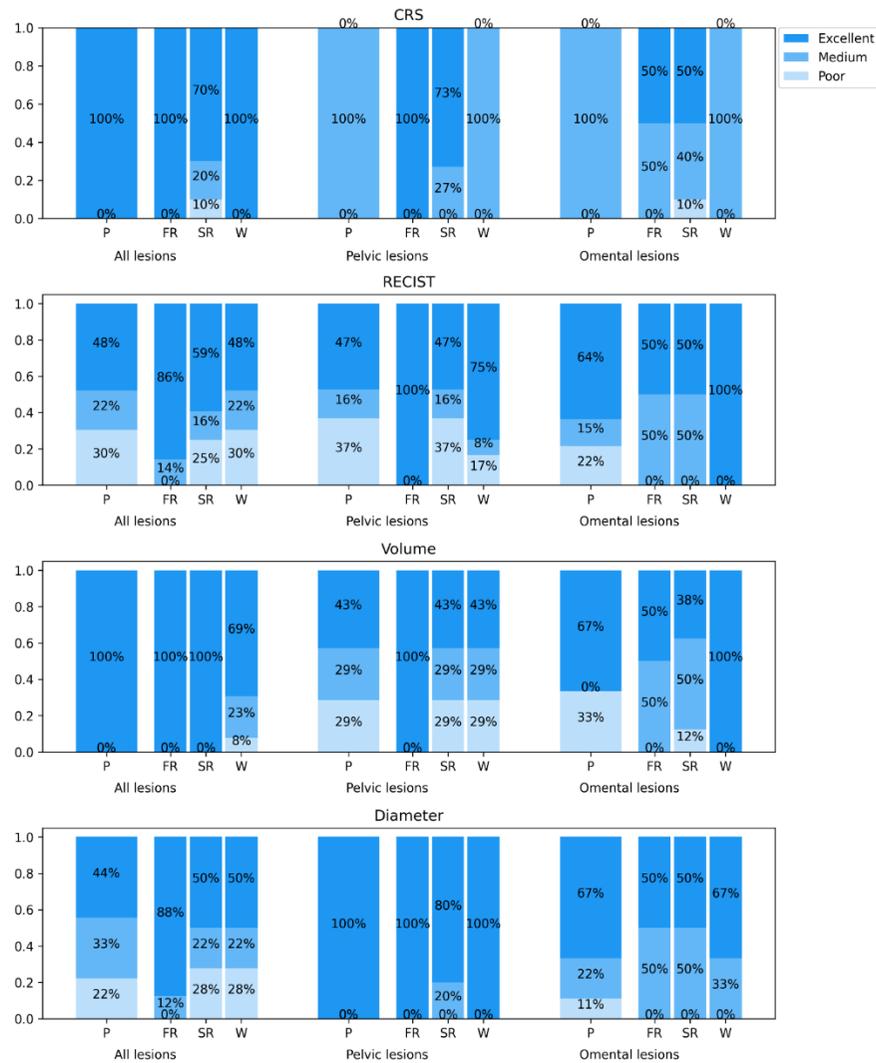

**Figure 5:** Comparison of ICC value distribution for Prediction of CRS, RECIST, Volume reduction and Diameter reduction (rows 1-4 respectively); separated into groups for All lesions, Pelvic lesions and Omental lesions. Stacked bars indicate percentage of features with Excellent (ICC>=0.9), Medium (0.7<=ICC<0.9) and Poor ICC values (ICC<0.7) for comparing PREDICTIVE (P) ICC value distributions with the PREDICITVE&ROBUST approaches Fully Robust (FR), Semi Robust (SR) and Weighted (W).